\documentclass[letterpaper]{article} 
\usepackage[submission]{aaai23}  
\usepackage{times}  
\usepackage{helvet}  
\usepackage{courier}  
\usepackage[hyphens]{url}  
\usepackage{graphicx} 
\usepackage{natbib}  
\usepackage{caption} 
\usepackage{url}
\usepackage{multirow}
\usepackage{amsmath}
\usepackage{amsfonts}
\usepackage{booktabs}
\usepackage{threeparttable}
\usepackage{nicefrac}
\usepackage{algorithm}
\usepackage{algorithmic}

\usepackage[justification=centering]{caption}
\usepackage{newfloat}
\usepackage{listings}
\DeclareCaptionStyle{ruled}{labelfont=normalfont,labelsep=colon,strut=off} 
\floatstyle{ruled}
\newfloat{listing}{tb}{lst}{}
\floatname{listing}{Listing}
\title{MA2GCN: Multi Adjacency relationship Attention Graph Convolutional Networks for Traffic
Prediction using Trajectory data}   
\author {
    Zhengke Sun*,\textsuperscript{\rm 1}
    Yuliang Ma, \textsuperscript{\rm 2}
}
\affiliations {
    \textsuperscript{\rm 1 2} Northeastern University, China\\
    zhengkesun@outlook.com, mayuliang@mail.neu.edu.cn
}

\usepackage{bibentry}

\begin{document}

\maketitle

\begin{abstract}

The problem of traffic congestion not only causes a large amount of economic losses, 
but also seriously endangers the urban environment. 
Predicting traffic congestion has important practical significance. 
So far, most studies have been based on historical data from sensors placed on different roads to predict future traffic flow and speed, 
to analyze the traffic congestion conditions of a certain road segment. 
However, due to the fixed position of sensors, it is difficult to mine new information. 
On the other hand, vehicle trajectory data is more flexible and can extract traffic information as needed. 
Therefore, we proposed a new traffic congestion prediction model - 
\underline{M}ulti \underline{A}djacency relationship \underline{A}ttention \underline{G}raph 
\underline{C}onvolutional \underline{N}etworks(MA2GCN). 
This model transformed vehicle trajectory data into graph structured data in grid form, 
and proposed a vehicle entry and exit matrix based on the mobility between different grids. 
At the same time, in order to improve the performance of the model, 
this paper also built a new adaptive adjacency matrix generation method and adjacency matrix attention module. 
This model mainly used gated temporal convolution and graph convolution to extract temporal and spatial information, respectively. 
Compared with multiple baselines, our model achieved the best performance on Shanghai taxi GPS trajectory dataset.
The code is available at \url{https://github.com/zachysun/Taxi_Traffic_Benchmark}.

\end{abstract}

\section{Introduction}

Intelligent transportation systems (ITS) is likely to become an integral part of the future smart cities and will include various services and applications, 
such as road traffic management, travel information system, public transport system management and autonomous vehicle.\cite{yuan2022machine}
Traffic prediction is an indispensable part of ITS and it is the process of analyzing and predicting key parameters (such as traffic flow, speed, etc.) 
to obtain future trends of traffic conditions. 

In the past, researchers often used statistical models to predict traffic flow, such as HA, ARIMA\cite{van1996combining}, 
SARIMA\cite{williams2003modeling}, etc. However, the above models cannot model nonlinear patterns. 
Later on, several classical machine learning methods was proposed, such as Support Vector Machine (SVM) \cite{hong2011traffic},
Markov Random Field (MRF) \cite{hoang2016fccf}, K-Nearest Neighbors (KNN) \cite{cai2016spatiotemporal}, etc. They can model nonlinear
and more complex data, but time-consuming feature engineering is necessary.
Due to the powerful learning ability, deep learning models have been widely used in traffic prediction in recent years and 
have achieved excellent performance on multiple datasets.
Some researchers regard a city area as $H\times W$-sized images and use Convolutional neural network (CNN) to extract spatial dependencies, 
such as spatiotemporal recurrent convolutional networks (SRCNs)\cite{yu2017spatiotemporal}, DeepST\cite{zhang2016dnn}, 
and ST-ResNet\cite{zhang2017deep}.
While, to capture the physical features and topological structure of the traffic network, 
many spatiotemporal models consider traffic network as spatial-temporal graphs and apply Graph convolutional network (GCN) to traffic prediction.
For example, T-GCN\cite{zhao2019t} uses Gated recurrent network (GRU) and GCN to extract temporal and spatial dependencies, respectively. 
DCRNN\cite{li2018diffusion} models traffic flow as a diffusion process and proposes diffusion convolution based on GCN. 
To capture the hidden spatial dependency in the data, Graph-WaveNet\cite{10.5555/3367243.3367303} develops a novel adaptive
adjecency matrix. Other works, for example, to learn the dynamic spatial-temporal correlations of traffic data, 
ASTGCN\cite{guo2019attention} applies attention mechanism into the spatio-temporal network module.
In addition, other related works have also applied attention mechanisms from the perspectives of hidden traffic parameters at 
different time steps\cite{bai2021a3t}, temporal patterns at different sampling periods\cite{D10.1145/3564754}, 
and graph convolutional neural networks of different layers\cite{Li9944937}. 

So far, traffic prediction problem is mainly based on sensor data(e.g.METR-LA, PEMS-BAY) or region in-out flow data
(e.g.TaxiBJ), which is fixed and difficult to reuse. However, few works start from trajectory data mining, 
building graph-structured data and doing traffic prediction.

\begin{figure}[ht]

    \center
    \includegraphics[scale=0.28]{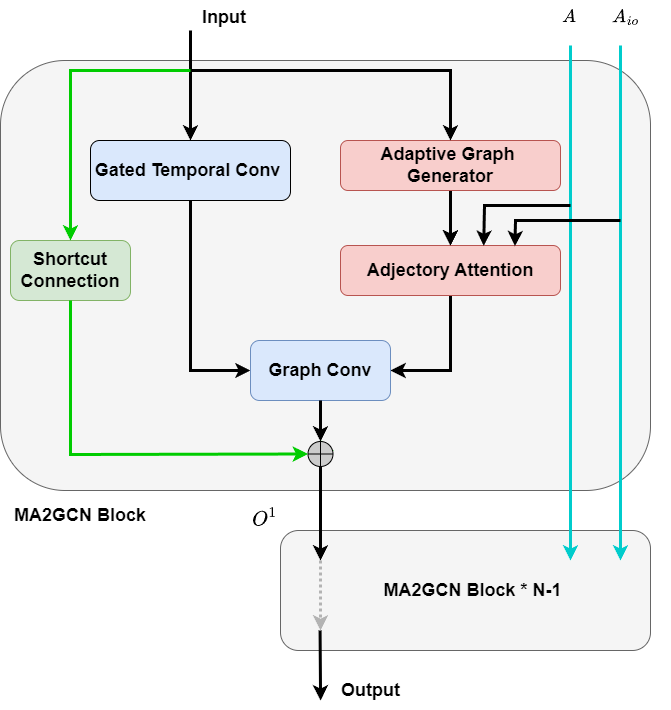}
    \caption{The framework of MA2GCN}
    \label{f1}

\end{figure}

The key contributions of our work as follows:
\begin{enumerate}
    \item[$\bullet$] We do not use datasets based on sensors or in-out flows, 
    but directly mine the trajectory data and transform it into graph-structured data for traffic prediction.
    \item[$\bullet$] We build a mobility adjacency matrix based on vehicle entry and exit data between grids. 
    Besides, we propose a multi adjacency relationship attention mechanism and a novel adaptive graph generator.
    \item[$\bullet$] Compared with multiple baselines, our proposed Multi Adjacency relationship Attention Graph Convolutional
    Networks(MA2GCN) (Fig.\ref{f1}.) achieved the best performance on the Shanghai taxi GPS trajectory dataset.
\end{enumerate}

\section{Methodology}

\subsection{Problem Definition}

In this study, we define $p_i=\left( x_i,y_i,t_i \right)$ as a spatio-temporal coordinate point, it means at time $t_i$, the position of the object is
$\left( x_i,y_i \right)$. 
The series $p_1\rightarrow \cdots \rightarrow p_i\rightarrow \cdots \rightarrow p_k$($k$ is total number of points) can be defined as a trajectory.
Besides, traffic grids can be seen as an undirected graph $G=\left( V, E, A \right)$, 
where $V$ is a set of N nodes(each grid corresponds to a node) and $E$ is a set of edges. 
$A\in \mathbb{R} ^{N\times N}$ denotes the adjacency matrix of graph $G$. The feature matrix of graph is denoted by $X\in R^{T\times N\times D}$, 
where T is the length of time series and D is the dimension of the features. Finally, given a graph $G$ and a feature matrix with P time steps, 
traffic prediction problem can be represented as Eq.\ref{eq:problem}.
\begin{equation}
\left\{ X_{\left( t-P \right) :t}, G \right\} \xrightarrow{f}\hat{X}_{t:\left( t+Q \right)}
\label{eq:problem}
\end{equation}
where $X_{\left( t-P \right) :t}=\left( X_{t-P}, X_{t-P+1},\cdots X_{t-1} \right) \in R^{P\times N\times D}$
and $\hat{X}_{t:\left( t+Q \right)}=\left( \hat{X}_t,\hat{X}_{t+1},\cdots \hat{X}_{t+Q-1} \right) \in R^{Q\times N\times D}$

\subsection{Graph Transformation}

We use grids with the same shape to divide the research area into $M * M$ parts. 
Then calculating the average instantaneous velocity of GPS points in each grid within a minimum time interval of 5 minutes.
Similarly, calculating the total traffic flows in each grid within 5 minutes. So we get two features of each grid.
And we build the initial adjacency matrix by treating all other grids around a grid as adjacent regions. Then, 
we propose a vehicle entry and exit matrix based on the mobility between different cars in different grids. 
This matrix represents the number and average travel time of taxis traveling from one grid to another over a period of time 
(usually optional 2-3 hours). Finally, We build the mobility adjacency matrix from the features of the vehicle entry and exit matrix.

\noindent \textbf{Vehicle Entry and Exit Matrix} This matrix can be defined as $E\in \mathbb{R} ^{S\times N\times N\times F}$,
where $S$ is the number of tme segments, the first $N$ denotes exit grids and the second $N$ denotes entry grids. $F$ represents features which including
traffic flows and average travel time. 
Given a taxi trajectory $\mathbb{P}$ and a grid $(m, n)$(at the $m^{th}$ row and $n^{th}$ column), if $p_i\notin \left( m,n \right)$ and
$p_{i+1}\in \left( m,n \right)$, we think this vehicle enter grid $(m, n)$. 
Similarly, if $p_j\notin \left( m',n' \right)$ and $p_{j+1}\in \left( m',n' \right)$, this vehicle enter grid $(m', n')$.
Thus, the travel time from grid $(m, n)$ to grid $(m', n')$ is defined as 
$p_{j+1}^{\left( t \right)}-p_{i+1}^{\left( t \right)}=t_{j+1}-t_{i+1}$.

\noindent \textbf{Mobility Adjacency Matrix} Generally speaking, the mobility between two places over a period of time is directly proportional to the traffic flow and 
inversely proportional to the average travel time. So the mobility adjacency matrix $A_{mo}$ can be defined as Eq.\ref{eq:mam}.
\begin{equation}
    A_{mo}^{ij}=\kappa \frac{\mu _{ij}}{\nu _{_{ij}}}
    \label{eq:mam}
\end{equation}
where $\mu _{ij}$, $\nu _{_{ij}}$ denote traffic flows and average traffic speed between the $i^{th}$ grid and the $j^{th}$ grid respectively,
and $\kappa$ is a hyperparameter.

\subsection{Spatio-Temporal Framework}

\noindent \textbf{Graph Convolution Layer}  Graph Convolutional Neural Network (GCN)\cite{kipf2017semisupervised} is a neural network model that 
applies convolutional operations to graph data. According to spectral graph theory, the laplace matrix of a graph is defined as $L=D-A$, 
where $A$ is the adjacency matrix, $D$ is the degree matrix, and is also a diagonal matrix $D_{ii}=\sum_j{A_{ij}}$. Symmetrically normalizing L to obtain
$L_{sym}=D^{-\frac{1}{2}}LD^{-\frac{1}{2}}$. 
$L$ is a positive semi-definite matrix, so we can perform an eigenvalue decomposition on L, 
obtaining $L=U\varLambda U^T$, where $\varLambda =\mathrm{diag}\left( \left[ \lambda _0,...,\lambda _{N-1} \right] \right) \in \mathbb{R} ^{N\times N}$ 
is a diagonal matrix and $U$ can be considered as the basis for a fourier transform, which is an orthogonal matrix.
We consider using convolution $g_{\theta}$ to process features $x$ of graph $G$ as Eq.\ref{eq:gconv}.
\begin{equation}
g_{\theta}\star _Gx=g_{\theta}\left( U\varLambda U^T \right) x=Ug_{\theta}\left( \varLambda \right) U^Tx
\label{eq:gconv}
\end{equation}
where $\star _G$ denotes a graph convolution operation. 

\noindent \textbf{Gated TCN}  In addition to RNN and LSTM, TCN is also commonly used for time series prediction due to its advantages 
such as fast speed and no gradient vanishing. 
The gating mechanism mainly controls the information flow of the network through some learnable parameters, 
and has important applications in recurrent networks.
The given input is $x\in \mathbb{R} ^{D\times T\times N}$, the representation of Gated TCN is as Eq.\ref{eq:gtcn}.
\begin{equation}
h=g\left( x*\theta _1+b_1 \right) \odot \sigma \left( x*\theta _2+b_2 \right)
\label{eq:gtcn}
\end{equation}
where $\theta _1$, $\theta _2$, $b_1$, $b_2$ is parameters, $g\left( \cdot \right)$ is an activation function,
$\sigma \left( \cdot \right)$ is the sigmoid function, and $\odot$ is the Hadamard product.

\subsection{Adaptive Graph Generator}
\begin{figure}[ht]

    \center
    \includegraphics[scale=0.35]{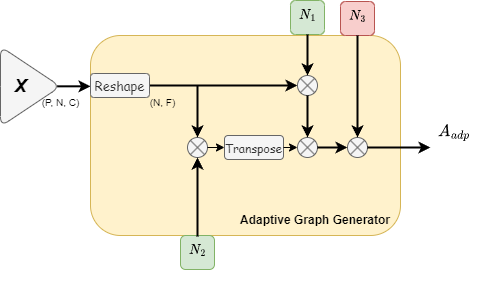}
    \caption{Adaptive Graph Generator}
    \label{f2}

\end{figure}

So far, many traffic prediction tasks used graphs based on predefined road structure information, such as the placement of sensors. 
These graphs are static during the training process and do not change, 
making them unable to represent dynamically changing road network spatial information.
In Graph-WaveNet\cite{10.5555/3367243.3367303}, an adaptive adjacency matrix based on random initialization was proposed,
but it didn't take into account the input features.
So we propose a novel adaptive graph generator(Fig.\ref{f2}.). Dynamic adjacency matrix is denoted by $A_{dy}\in \mathbb{R} ^{N\times N}$.
We define $\tilde{X}\in \mathbb{R} ^{N\times F}$, where $F=P\times D$. 
Three learnable weight matrices are denoted by $N_1\in \mathbb{R} ^{F\times H}$, 
$N_2\in \mathbb{R} ^{F\times H}$, and $N_3\in \mathbb{R} ^{N\times N}$, respectively, where $H$ is the 
hidden dimension. Therefore, the adaptive graph generator is represented as Eq.\ref{eq:dgg}.
\begin{equation}
A_{dy}=SoftMax\left( \left( \tilde{X}N_1\left( \tilde{X}N_2 \right) .T \right) N_3 \right)
\label{eq:dgg}
\end{equation}

\subsection{Multi Adjacency relationship Attention}
\begin{figure}[h]

    \center
    \includegraphics[scale=0.3]{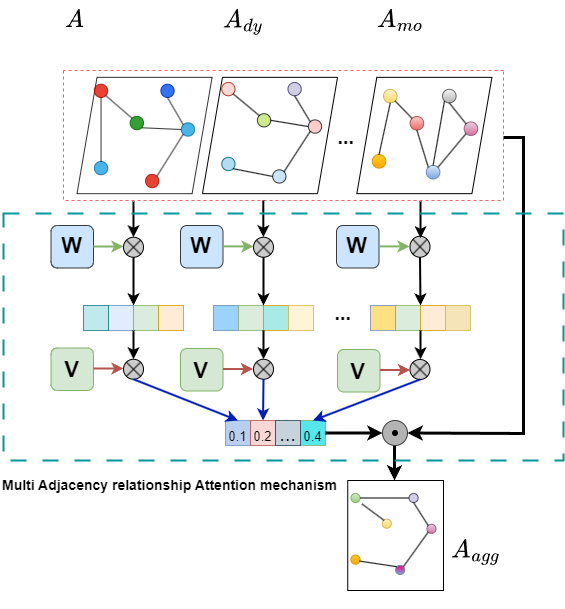}
    \caption{Multi Adjacency relationship Attention mechanism}
    \label{f3}

\end{figure}

We propose a Multi Adjacency relationship Attention Mechanism(Fig.\ref{f3}.). 
This attention mechanism models different types of adjacency matrices, 
with the goal of enabling the model to focus on more important adjacency matrices, 
thereby more accurately learning the spatial information of the input.

Our model selects four types of adjacency matrices, namely the initial adjacency matrix$A$, 
the square of the initial adjacency matrix$A^2$, the dynamic adjacency matrix$A_{dy}$, 
and the mobility adjacency matrix$A_{mo}$. 

\begin{algorithm}[h]
    \caption{Multi Adjacency relationship Attention Steps}
    \label{ma2}
    \begin{algorithmic}[1]
    \REQUIRE The set of adjacency matrices $\mathbb{A}$
    \ENSURE The aggregated adjacency matrix $A_{agg}$
    
    \FOR{$A_i\in\mathbb{A}$ \COMMENT{$i\in \left[ 1,4 \right]$}} 
        \STATE Linear layer-Activation function: $x=ReLU\left( A_iW \right)$
        \STATE Linear kayer: $x\prime=xV$
        \STATE Average: $\alpha _i=\frac{\sum_{j=0}^N{x\prime_j}}{N}$
    
    \ENDFOR
    \STATE Normalization: $\tilde{\alpha}_i=\frac{exp\left( \tilde{\alpha}_i \right)}{\sum_{k=1}^4{exp\left( \tilde{\alpha}_k \right)}}$
    \STATE Aggregation: $A_{agg}=\tilde{\alpha}_1A+\tilde{\alpha}_2A^2+\tilde{\alpha}_3A_{dy}+\tilde{\alpha}_4A_{mo}$
    
    \end{algorithmic}
    \end{algorithm}

Define a set of adjacency matrices consisting of above adjacency matrices $\mathbb{A} =\left[ A,A^2,A_{dy},A_{mo} \right]$.
Define the output as the aggregated adjacency matrix $A_{agg}$, as well as two weight matrices $W\in \mathbb{R} ^{N\times H'}$,
$V\in \mathbb{R} ^{H'\times 1}$($H'$ is the hidden dimension). The steps are as Alg.\ref{ma2}.

\section{Experiment}

All experiments in this section were implemented using Pytorch 1.12.1 in Python 3.8.0 on a server
with AMD Ryzen 9 6900HX @ 3.30GHz and one Nvidia Geforce RTX 3070 Ti Laptop GPU.

\subsection{Dataset}
We validate our model on Shanghai taxi GPS trajectory dataset. The dataset was collected by Smart City Research Group of 
HKUST. 
It described the GPS trajectory of a total of 4316 Shanghai taxis on February 20, 2007. 
The information of each taxi was stored in a document in txt format, with 7 fields in each row, 
namely taxi ID, timestamp, logitude, latitude, instantaneous speed, angle from the north in clockwise direction and 
current status of this taxi.

\subsection{Settings}

This experiment sets the initial learning rate to 0.001 and the weight decay rate to 0.0001;
The batch size is 8 and the number of training epochs is 100; 
The ratio of training set, validation set, and test set is 7:2:1;
The total number $N$ of grids after division is 225 (side length is 15);
The input sequence duration $P$ is 1 hour (12 time steps), and the output sequence duration $Q$ is 5 minutes (1 time step);
The order $K$ of the Chebyshev polynomial in graph convolution is set to 3; 
The hidden dimension $H\&H'$ are set to 128; The number of MA2GCN block is set to 3.
Mean absolute error (MAE), mean absolute percentage error (MAPE), 
root mean square error (RMSE) are used as the evaluation metrics.

We also use MAE to build loss function, which is as Eq.\ref{eq:loss}.

\begin{equation}
    loss=\theta *l\left( y_{pred}^{\left( 1 \right)},y_{true}^{\left( 1 \right)} \right) +\left( 1-\theta \right) *l\left( y_{pred}^{\left( 2 \right)},y_{true}^{\left( 2 \right)} \right)
    \label{eq:loss}
\end{equation}
where $y_{pred}^{\left( 1 \right)}$ and $y_{pred}^{\left( 2 \right)}$ are predicted values of speed and traffic flows, respectly;
$y_{true}^{\left( 1 \right)}$ and $y_{true}^{\left( 2 \right)}$ are true values of speed and traffic flows, respectly;
$\theta$ is a hyper-parameter, which is set to 0.5.

\subsection{Baselines}

\begin{enumerate}
    \item[$\bullet$] \textbf{ARIMA}:\cite{kumar2015short}
    it is a classical time series analysis method used to forecast or explain time series data.
    \item[$\bullet$] \textbf{SVR}:\cite{10.1016/j.eswa.2008.07.069}
    it is a regression model used to fit and predict time series data.
    \item[$\bullet$] \textbf{LSTM}:\cite{hochreiter1997long}
    it is a variant of recurrent neural networks (RNN) commonly used for time series analysis.
    \item[$\bullet$] \textbf{STGCN}:\cite{ijcai2018p505}
    use GCN to model spatial dependencies and use temporal gated convolution to model temporal dependencies.
    \item[$\bullet$] \textbf{ASTGCN}:\cite{guo2019attention}
    use a spatial-temporal attention mechanism to learn the dynamic spatial-temporal correlations.
    \item[$\bullet$] \textbf{Graph-WaveNet}:\cite{10.5555/3367243.3367303}
    use an adaptive adjacency matrix to learn dynamic traffic process.
\end{enumerate}

\subsection{Performance Comparison}

In order to verify the performance of our model, we selects models based on statistical learning, 
machine learning, and deep learning for comparison experiments.

\begin{table}[h]
    \centering
    \caption{Model comparison on Shanghai taxi GPS trajectory dataset in terms of MAE, MAPE and RMSE} 
    \label{t:per}
    \begin{tabular}{c|ccc}
        \toprule
        Model & MAE & MAPE & RMSE\\
        \midrule
        ARIMA & 8.86 & 0.54 & 16.82\\
        SVR & 16.12 & 0.71 & 49.07\\
        LSTM & \underline{7.95} & \underline{0.47} & \underline{15.79}\\
        STGCN & 12.73 & 0.81 & 25.50\\
        ASTGCN & 9.33 & 0.56 & 18.63\\
        Graph-WaveNet & 10.00 & 0.62 & 21.82\\
        MA2GCN(Ours) & \textbf{7.47} & \textbf{0.44} & \textbf{14.99}\\
        \bottomrule
    \end{tabular}
\end{table}

Our model achieves optimal performance on all three metrics, 
with improvements of 6.0\%, 6.4\%, and 5.1\% respectively.(Tab.\ref{t:per}.)

\subsection{Ablation Study}

We also conduct ablation experiments to obtain the contribution of each module to the overall model.
To simplify, 'att' represents the multi adjacency relationship attention mechanism, 'dy' represents the
adaptive graph generator. 
In order to eliminate the influence of the attention mechanism, 
we assigns the same weight to four adjacency matrices, all of which are 0.25;
In order to eliminate the influence of dynamic graph generator, we only includes the remaining three types of matrices 
in the adjacency matrix list; 
In order to simultaneously eliminate the impact of both, 
we ensures that the adjacency matrix of GCN is always the initial adjacency matrix.

\begin{table}[h]
    \centering
    \caption{Ablation study on Multi Adjacency relationship Attention and Adaptive Graph Generator} 
    \label{t:abl}
    \begin{tabular}{c|ccc}
        \toprule
        Model & MAE & MAPE & RMSE\\
        \midrule
        MA2GCN(w/o att \& dy) & 7.90 & \underline{0.52} & 15.40\\
        MA2GCN(w/o att) & 7.95 & 0.56 & 15.68\\
        MA2GCN(w/o dy) & \underline{7.84} & \underline{0.52} & \underline{15.36}\\
        MA2GCN & \textbf{7.47} & \textbf{0.44} & \textbf{14.99}\\
        \bottomrule
    \end{tabular}
\end{table}

Both modules make positive contributions to model performance, 
with both improving prediction performance by 6.0\% and 4.7\%, respectively.(Tab.\ref{t:abl}.)

\section{Conclusion}

In this paper, we propose a novel model called MA2GCN to predict both traffic speed and flows. Specifically, 
we transform the trajectory data to the graph-structured data and develop a mobility adjacency matrix. 
We propose the multi adjacency relationship attention mechanism, 
which is a novel attention module and is used to learn the correlations of different adjacency matrices.
Besides, we use dynamic graph generator to capture complex traffic process.
Experiments on a real-world dataset show that the prediction accuracy of our proposed model is better than existing
models. Future work can focus on adaptive grid partitioning and mining more hidden connections in trajectories.

\bibliography{aaai23}
\end{document}